%% file: main.tex
\documentclass[letterpaper, 10 pt, conference]{ieeeconf}  % Comment this line out if you need a4paper

\usepackage{amsmath, amssymb}
\usepackage{cite}

\usepackage{float}
\usepackage{graphicx}
\usepackage{tikz}
\usepackage{subcaption}
\usepackage{pifont}
\usepackage{siunitx}
\DeclareMathAlphabet{\pazocal}{OMS}{zplm}{m}{n}
\IEEEoverridecommandlockouts                              % This command is only needed if 
\title{\LARGE \bf
Depth-constrained ASV navigation with deep RL and limited sensing
}

\author{Amirhossein Zhalehmehrabi, Daniele Meli, Francesco Dal Santo, Francesco Trotti and Alessandro Farinelli% <-this % stops a space
\thanks{*The authors are with the Department of Computer Science, University of Verona, Italy.}% <-this % stops a space
}

\begin{document}

\maketitle
\thispagestyle{empty}
\pagestyle{empty}

%%%%%%%%%%%%%%%%%%%%%%%%%%%%%%%%%%%%%%%%%%%%%%%%%%%%%%%%%%%%%%%%%%%%%%%%%%%%%%%%
\begin{abstract}

Autonomous Surface Vehicles (ASVs) play a crucial role in maritime operations, yet their navigation in shallow-water environments remains challenging due to dynamic disturbances and depth constraints. Traditional navigation strategies struggle with limited sensor information, making safe and efficient operation difficult. In this paper, we propose a reinforcement learning (RL) framework for ASV navigation under depth constraints, where the vehicle must reach a target while avoiding unsafe areas with only a single depth measurement per timestep from a downward-facing Single Beam Echosounder (SBES). To enhance environmental awareness, we integrate Gaussian Process (GP) regression into the RL framework, enabling the agent to progressively estimate a bathymetric depth map from sparse sonar readings. This approach improves decision-making by providing a richer representation of the environment. Furthermore, we demonstrate effective sim-to-real transfer, ensuring that trained policies generalize well to real-world aquatic conditions. Experimental results validate our method’s capability to improve ASV navigation performance while maintaining safety in challenging shallow-water environments.

\end{abstract}

%%%%%%%%%%%%%%%%%%%%%%%%%%%%%%%%%%%%%%%%%%%%%%%%%%%%%%%%%%%%%%%%%%%%%%%%%%%%%%%%
\input{sections/1.introduction}
\input{sections/2.related_works}
\input{sections/3.method}

\input{sections/4.simulation_experiments}
\input{sections/5.Conclusion}
%\section{INTRODUCTION}

%This template provides authors with most of the formatting specifications needed for preparing electronic versions of their papers. All standard paper components have been specified for three reasons: (1) ease of use when formatting individual papers, (2) automatic compliance to electronic requirements that facilitate the concurrent or later production of electronic products, and (3) conformity of style throughout a conference proceedings. Margins, column widths, line spacing, and type styles are built-in; examples of the type styles are provided throughout this document and are identified in italic type, within parentheses, following the example. Some components, such as multi-leveled equations, graphics, and tables are not prescribed, although the various table text styles are provided. The formatter will need to create these components, incorporating the applicable criteria that follow.

\addtolength{\textheight}{-12cm}   % This command serves to balance the column lengths

%%%%%%%%%%%%%%%%%%%%%%%%%%%%%%%%%%%%%%%%%%%%%%%%%%%%%%%%%%%%%%%%%%%%%%%%%%%%%%%%

%%%%%%%%%%%%%%%%%%%%%%%%%%%%%%%%%%%%%%%%%%%%%%%%%%%%%%%%%%%%%%%%%%%%%%%%%%%%%%%%

%%%%%%%%%%%%%%%%%%%%%%%%%%%%%%%%%%%%%%%%%%%%%%%%%%%%%%%%%%%%%%%%%%%%%%%%%%%%%%%%
\section*{ACKNOWLEDGMENT}

The work was carried out within the Interconnected Nord-Est Innovation Ecosystem (iNEST) and received funding from the European Union Next-GenerationEU (PIANO NAZIONALE DI RIPRESA E RESILIENZA (PNRR) – MISSIONE 4 COMPONENTE 2, INVESTIMENTO 1.5 – D.D. 1058
23/06/2022, ECS00000043).

\bibliographystyle{ieeetr}  % or another style like plain, unsrt, apalike, etc.
\bibliography{biblio.bib}

\end{document}

%% file: sections/1.introduction.tex
\section{INTRODUCTION}
Autonomous Surface Vehicles (ASVs) are unmanned vessels increasingly employed for a variety of maritime operations, including environmental monitoring, search-and-rescue, and surveillance. However, operating in aquatic environments presents significant challenges due to the unpredictable and dynamic nature of water. Factors such as water currents, waves, and other external disturbances constantly affect an ASV’s motion, complicating both precise navigation and the implementation of effective motion compensation and control strategies \cite{wu2024overview, hodges2023evaluation}.

In addition to these challenges, safe navigation becomes even more critical in shallow-water scenarios, where depth constraints must be strictly observed. Recent advancements in shallow-water bathymetric surveying have highlighted the importance of millimeter-level positioning accuracy, particularly in coastal regions with depths below 2.0 m, where exposed rock formations and artificial structures create complex navigation environments \cite{hyun2023bathymetric, wilson2018adaptive}.

Moreover, the wave-following behavior of ASVs causes rapid shifts, rolls, and pitches, intensifying navigation challenges and increasing the risk of drifting into deep waters or colliding with the shoreline. Such occurrences may lead to mission failure, equipment damage, or loss of communication, especially in applications such as environmental monitoring and seabed mapping. Therefore, addressing these operational challenges necessitates the development of advanced control systems capable of adapting to the continuously changing environmental conditions, ensuring both safe and efficient ASV operations.

In recent years, deep RL has shown promise in robotics for addressing complex challenges \cite{lee2020learning,hwangbo2017control, gu2017deep}. Model-free policy search specifically develops a parametric policy through direct interactions with the environment, making it particularly effective for complex problems that are difficult to model. Additionally, this approach facilitates the optimization of high-capacity neural network policies, which can accommodate various forms of state representations as inputs, thereby providing considerable flexibility in algorithm design.

\begin{figure}
    \centering
    \includegraphics[width=\linewidth]{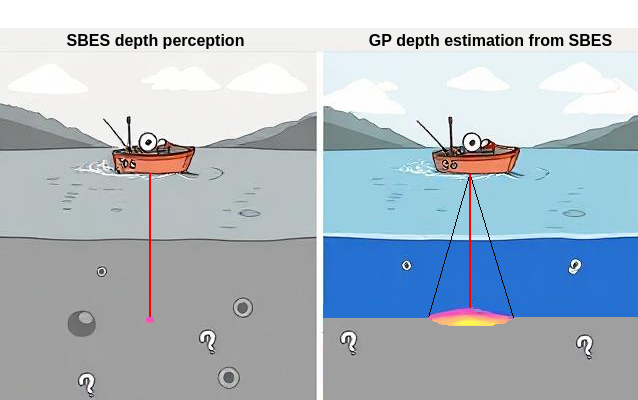}
    \vspace{-15pt}
    \caption{Illustration of the improved observation space achieved through GP.}
    \vspace{-10pt}
    \label{fig:intro}
\end{figure}

In this paper, we address the problem of ASV navigation under depth constraints using RL, where the vehicle must reach a target while avoiding both excessively deep areas and the shoreline, so called \textit{unsafe areas}. Specifically, we consider the practical case where limited sensing devices are available on the ASV, using a single downward-facing Single Beam Echosounder (SBES) that provides a single depth measurement at the current position. This constraint reflects a growing trend toward low-cost, lightweight, and easily deployable ASVs for environmental monitoring applications, where minimal sensor setups are preferred to reduce complexity and cost, such as in the INTCATCH project, which promotes portable and affordable solutions for water quality assessment \footnote{INTCATCH Project – Development and application of Novel, Integrated Tools for monitoring and managing Catchments. https://www.intcatch.eu}. This indirect information requires the agent to infer the locations of unsafe areas without explicit knowledge of their positions. This assumption makes the environment partially observable and significantly limits the agent's awareness of the surrounding terrain, preventing the agent from inferring a safe direction of motion and thus making the aquatic navigation problem very challenging. Partial observability is commonly addressed through four main approaches: belief state methods \cite{thrun2002probabilistic}, which maintain a distribution over possible states; history-based methods \cite{hausknecht2015deep}, such as LSTMs, that use past observations; model-based methods \cite{deisenroth2011pilco}, which learn and plan with environment models; and direct observation methods \cite{levine2016end}, which operate solely on current inputs. Our approach frames partial observability as a belief-state estimation problem, and we employ Gaussian Process (GP) regression to incrementally estimate a spatial belief of the bathymetric map from sparse SBES measurements (Figure~\ref{fig:intro}). While GPs offer principled uncertainty estimates, several studies have shown that predictive variance can saturate in regions with sparse data due to limitations in kernel design and finite-rank approximations \cite{quinonero2005analysis, sanz2025gaussian, guan2021measuring}. This leads to overconfident predictions that blur the distinction between observed and inferred areas, compromising the reliability of uncertainty quantification.

To overcome this, we build on local and sparse GP approaches \cite{nguyen2008local, snelson2005sparse}, introducing an additional confidence mechanism that explicitly mitigates variance shrinkage in unobserved regions. The resulting belief map, updated in real time, is embedded in the agent’s observation space and significantly improves policy learning and transfer.

\begin{figure}
    \centering
    \includegraphics[width=\linewidth]{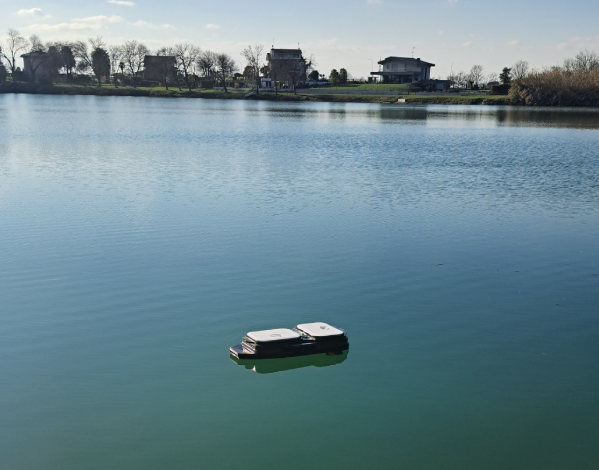}
    \vspace{-15pt}
    \caption{An image showing the real ASV used for real-world experiments.}
    \vspace{-10pt}
    \label{fig:real_asv}
\end{figure}

We validate our method through sim-to-real deployment on a custom-built ASV equipped with only a SBES, with no additional fine-tuning required in the real world (Figure~\ref{fig:real_asv}).
%Our approach relies on indirect information—a single depth reading per timestep—. This makes the task significantly more challenging, as the available information is insufficient for straightforward safe navigation.
The main contributions of this paper are the following:
\begin{itemize}
\item A spatial belief model based on localized Gaussian Process regression is introduced, enabling uncertainty-aware terrain estimation from sparse SBES readings.

\item A novel confidence-based uncertainty quantification mechanism is proposed to mitigate variance saturation in GP models.

\item The first integration of real-time spatial belief updates into an RL policy for depth-constrained ASV navigation is presented.

\item Effective zero-shot sim-to-real transfer is demonstrated, with the trained policy deployed on a real ASV without additional fine-tuning.

\item Code and simulation tools are going to be released to promote reproducibility and support further research in reinforcement learning for robotics.

\end{itemize}

%% file: sections/2.related_works.tex
\section{RELATED WORKS}

%The challenge of bathymetry-aware navigation can be conceptualized as an obstacle avoidance problem, where unsafe areas represent obstacles that must be avoided \textbf{without direct distance measurements}. 

Current research in the domain of aquatic navigation, particularly surface navigation using ASVs, generally falls into three primary categories: Model-based methods, RL-based methods and depth-constrained navigation approaches.
\subsection{Model-based Aquatic Navigation}
Model-based navigation approaches, such as Model Predictive Control, leverage a system dynamics model to predict future states and optimize control actions accordingly. These methods have been widely used in ASV navigation due to their ability to handle constraints and provide stable, interpretable control; however, they often struggle with modeling inaccuracies and computational complexity in highly dynamic environments \cite{xue2021real,menges2024nonlinear,zheng2014trajectory}. 
Unlike model-based approaches, RL does not rely on an explicit dynamics model, allowing it to handle unmodeled effects and adapt to complex environments. Following recent trends in complex domains such as UAV control \cite{song2023reaching} we focus on RL, which has been shown to outperform optimal control by optimizing a better objective, leading to more robust control strategies.
\subsection{Reinforcement Learning for Aquatic Navigation}
A fundamental challenge in navigation is obstacle avoidance. In our case, the objective is to avoid unsafe areas, such as shallow regions, which act as obstacles. However, unlike traditional settings, we do not have access to direct distance measurements to these hazards—for example, no side-scan or multi-beam sonar—making the task more challenging. 

Several studies have applied RL to collision avoidance and path planning in aquatic environments. Zhou et al.~\cite{zhou2019learn} developed RL algorithms for both single and multi-ASV path planning with a focus on obstacle avoidance, though their approach assumed perfect knowledge of obstacle locations within a fully observable simulated environment. Meyer et al.~\cite{meyer2020taming} used rangefinder-based perception systems to train RL agents to avoid static obstacles. Other studies have also addressed obstacle avoidance \cite{meyer2020colreg, zhao2019control, larsen2021comparing, zhang2021model}, but all incorporate explicit obstacle information through sensors such as rangefinders or positional data.  

The key distinction of our work is that it relies solely on a single SBES, which provides depth readings only at the ASV's current location. We construct a probabilistic model to infer the surrounding areas, whereas existing RL methods for aquatic navigation use sensing mechanisms that directly detect obstacles. %While both our approach and prior RL methods operate under partial observability, our approach utilizes an indirect source of information, unlike existing methods that explicitly sense obstacles.  

\subsection{Depth-Constrained Navigation}

Research focusing specifically on depth-constrained navigation has primarily utilized waypoint-based strategies \cite{hyun2023bathymetric} and adaptive control mechanisms \cite{wilson2018adaptive} to ensure safe operation in varying underwater topographies. These methods rely on predefined paths or reactive control adjustments based on bathymetric data.
Our research bridges these domains by employing RL techniques to solve the depth-constrained navigation problem using insufficient sensory input. By training an RL agent to effectively interpret limited depth information, we enable autonomous navigation in shallow-water environments without requiring comprehensive bathymetric maps or multiple sensing modalities.

\subsection{Uncertainty Estimation in Gaussian Processes}

GPs are commonly used for spatial inference and uncertainty estimation in robotics and environmental monitoring \cite{rasmussen2003gaussian}. However, a known limitation is variance saturation in regions with sparse data, caused by limited kernel expressiveness and low-rank approximations \cite{quinonero2005analysis, sanz2025gaussian, guan2021measuring}. This results in overconfident predictions that fail to distinguish observed from inferred regions.
To address this, we build on localized and sparse GP approaches \cite{nguyen2008local, snelson2005sparse} and introduce a covariance-based confidence proxy that better captures observational density. This improves the quality of the belief map and enhances policy learning under partial observability.

%% file: sections/3.method.tex
\section{METHOD}
We now introduce the depth-constrained ASV navigation problem and outline our RL- and GP-based solution.
\subsection{Task Definition}
\begin{figure}
    \centering
    \begin{tikzpicture}
        \node at (0,0) {\includegraphics[width=0.24\textwidth]{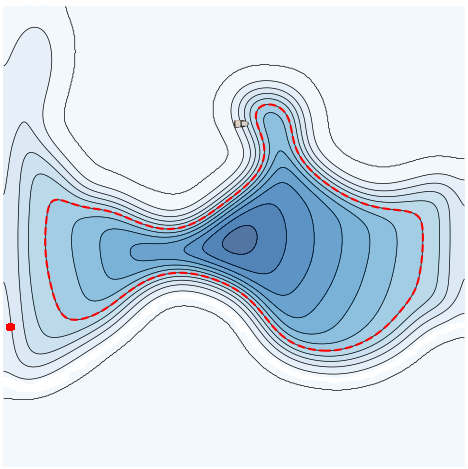}}; % Main image
        \draw[red, thick] (-0.35,0.5) rectangle (0.75,1.6); % Highlighted area
        \node at (4,0) {\includegraphics[width=0.24\textwidth]{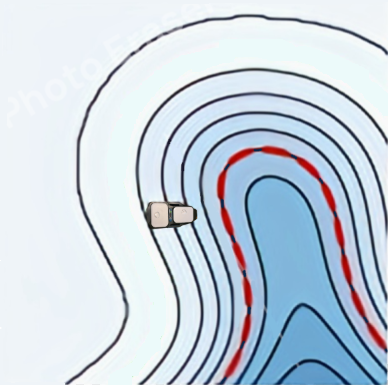}}; % Zoomed image
        \draw[red, thick, ->] (0.75,1.05) -- (2.5,0); % Arrow pointing to zoomed part
    \end{tikzpicture}
    \vspace{-10pt}
    \caption{Example of generated depth maps. The red dot marks the target point, while the red dashed line indicates the \( L_d \) level set contour, defining the unsafe area. The outermost contour represents the shoreline. The zoomed-in section shows the ASV navigating near shallow waters.}
    \vspace{-10pt}
    \label{fig:depth_maps}
\end{figure}

The depth-constrained ASV navigation task (Figure \ref{fig:depth_maps}) is formulated as an optimization problem, aiming to minimize the distance to a designated target while avoiding unsafe areas, represented by the shoreline and the regions where the water depth exceeds a specified threshold \( L_d \).
The ASV relies only on a downward-facing SBES, which helps detect the shoreline based on depth variations. 

Our goal is to develop a generalizable strategy for solving this task in unknown water basins, starting from a RL policy learned in arbitrary simulation environments.
To this aim, we use online GP regression to estimate the depth map of the basin, as the ASV navigates and collects readings from the SBES. We then define the navigation problem as a POMDP, with an additional observation given by the GP estimation.
% For the simulation phase, we used the dynamic model of the ASV explained in \ref{method.dynamics} to create the training environment. 

\subsection{ASV Dynamics}
\label{method.dynamics}

Following the formulation in \cite{gierusz2016modelling}, the ASV is modeled as a six-degree-of-freedom (6 DOF) rigid body system with mass $m$. The full nonlinear dynamics are governed by the following equations:
\begin{equation}
    \begin{aligned}
    m(\dot{u} + qw - rv) &= X_{\text{tot}}, & I_x \dot{p} + (I_z - I_y)qr &= K_{\text{tot}} \\
    m(\dot{v} + ru - pw) &= Y_{\text{tot}}, & I_y \dot{q} + (I_x - I_z)rp &= M_{\text{tot}} \\
    m(\dot{w} + pv - qu) &= Z_{\text{tot}}, & I_z \dot{r} + (I_y - I_x)pq &= N_{\text{tot}}
    \end{aligned}
\end{equation}
where \( X_{\text{tot}} \), \( Y_{\text{tot}} \), and \( Z_{\text{tot}} \) represent the resultant forces in surge (\( x \)), sway (\( y \)), and heave (\( z \)) directions respectively, while \( K_{\text{tot}} \), \( M_{\text{tot}} \), and \( N_{\text{tot}} \) denote the moments about the roll (\( x \)), pitch (\( y \)), and yaw (\( z \)) axes. The terms \( I_x \), \( I_y \), and \( I_z \) correspond to the principal moments of inertia, with \( u \), \( v \), \( w \) and \( p \), \( q \), \( r \) representing the linear and angular velocities in the body-fixed frame.

\subsection{GP depth estimation}
Incorporating only a single depth measurement per time step provides insufficient information about the underlying depth map, making safe navigation challenging, as evidenced by the poor performance of our baseline model w/o GP in the empirical evaluation. To address this limitation, we employ localized GP regression to estimate the depth of nearby locations (Figure \ref{fig:intro}), leveraging past SBES readings to construct a probabilistic representation of the underwater terrain. This approach enables the vehicle to infer depth variations beyond its immediate measurement, improving decision-making in sparse sensing conditions. 

Let \( d: \mathbb{R}^2 \rightarrow \mathbb{R} \) denote the continuous depth function of the underwater terrain. At time \( t \), the ASV records a sonar measurement \( z_t \) corresponding to the depth \( d(\mathbf{x}_t) \) at its current location \( \mathbf{x}_t \in \mathbb{R}^2 \), given by  
\begin{equation}
z_t = d(\mathbf{x}_t) + \epsilon, \quad \epsilon \sim \mathcal{N}(0, \sigma_{\rm SBES}^2).
\end{equation}
where $\epsilon$ is the measurement noise and $\sigma_{\rm SBES}^2$ is the sensor's variance.
To maintain computational efficiency and integrate seamlessly with the RL pipeline, we represent the spatial belief over depth as a discrete 2D grid, where each cell corresponds to a square area of \( \xi \times \xi \) cm\(^2\). That is, the continuous domain is discretized into a regular grid, and GP estimates are maintained at the center of each grid cell.
Let $\mathcal{M}(\mathbf{x}) = \left\{ \mathbf{x'} \mid k(\mathbf{x}, \mathbf{x'}) > \delta \right\}$\footnote{In our experiments, we set $\delta=0.01$ based on empirical observations.} denote a local neighborhood of \( \mathbf{x} \), where $k$ is the covariance function.
 Within this region, we assume a GP prior:  
\begin{equation}
    d(\mathbf{x}) \sim \mathcal{GP}(\mu_0(\mathbf{x}), k(\mathbf{x}, \mathbf{x}')), \\
\end{equation}
where \( \mu_0 (\mathbf{x}) \) is the prior mean function. 
For each location $\mathbf{x}^i_t \in \mathcal{M}(\mathbf{x}_t)$, the posterior parameters of the Gaussian process are updated by incorporating the new data in a Bayesian manner.
\begin{equation}
\label{eq:GP}
\begin{aligned}
\sigma_{t+1}^2(\mathbf{x}^i_t) = \frac{\sigma_t^2(\mathbf{x}^i_t) \cdot \sigma_{{\rm w}}^2}{\sigma_t^2(\mathbf{x}^i_t) + \sigma_{{\rm w}}^2} \\
\mu_{t+1}(\mathbf{x}^i_t) = \frac{\sigma_{{\rm w}}^2 \cdot \mu_t(\mathbf{x}^i_t) + \sigma_t^2(\mathbf{x}^i_t) \cdot z_t}{\sigma_t^2(\mathbf{x}^i_t) + \sigma_{{\rm w}}^2}
\end{aligned}
\end{equation}
where 
\begin{equation*}
\sigma_{{\rm w}}^2 = \frac{\sigma_{{\rm SBES}}^2}{k(\mathbf{x}^i_t, \mathbf{x}_t)^2}
\end{equation*}
represents the sensor variance weighted by the spatial relationship function.
The update mechanism derives from the Gaussian conjugate prior framework, ensuring that new data incorporation leads to tractable sequential updates. The resulting posterior distribution reflects both prior knowledge and the localized impact of the new observation.

To mitigate variance saturation in regions with sparse data, we define a confidence proxy \( C \) that leverages the GP covariance structure to better distinguish between observed and inferred locations. This proxy provides a more reliable uncertainty signal for downstream decision-making.

For each estimated location \( \mathbf{x}^i_t \), the confidence proxy is defined as:  
\begin{equation}
\label{eq:proxy}
C(\mathbf{x}^i_t) = \frac{\sum_{\mathbf{x}^j \in \mathcal{O}} k(\mathbf{x}^i_t, \mathbf{x}^j)}{\sum_{\mathbf{x}^k \in \mathbb{R}^2} k(\mathbf{x}^i_t, \mathbf{x}^k)} = \frac{\sum_{\mathbf{x}^j \in \mathcal{O}} k(\mathbf{x}^i_t, \mathbf{x}^j)}{\sum_{\mathbf{x}^k \in \mathcal{M}} k(\mathbf{x}^i_t, \mathbf{x}^k)}
\end{equation}
where \( \mathcal{O} \) represents the set of previously observed locations. The denominator ensures normalization by considering the total covariance sum over all spatial locations, restricting \( C(\mathbf{x}^i_t) \) to the range \([0,1]\). Furthermore, for every observed point \( \mathbf{x}_t \), the confidence proxy is set to \( C(\mathbf{x}_t) = 1 \), ensuring maximum confidence.

Once the update is applied, the measurement \( (\mathbf{x}_t, z_t) \) is discarded and does not contribute to future updates. This localized GP update allows the ASV to incrementally refine its depth estimates in real-time while maintaining computational efficiency.

\subsection{Gradient-Based Extrapolation}
\label{future_update}
As the boat moves along a trajectory \(\{\mathbf{x}_t\}_{t=0}^{T}\), it collects depth measurements sequentially. Assuming that \(d\) is differentiable, a first-order Taylor expansion about \(\mathbf{x}_t\) for a small displacement \(\Delta \mathbf{x}\) yields
\begin{equation}
    d(\mathbf{x}_t + \Delta \mathbf{x}) \approx d(\mathbf{x}_t) + \nabla d(\mathbf{x}_t)^\top \Delta \mathbf{x}.
\end{equation}

If the boat is moving in the direction of the unit vector $\mathbf{u}_t$ and we wish to estimate the depth at a new point \(\mathbf{x}_t + \Delta s\, \mathbf{u}_t\) (with \(\Delta s > 0\)), then
\begin{equation}
    d(\mathbf{x}_t + \Delta s\, \mathbf{u}_t) \approx d(\mathbf{x}_t) + \Delta s \, (\nabla d(\mathbf{x}_t)^\top \mathbf{u}_t).
\end{equation}

We approximate the directional derivative using finite differences, Thus, the estimated depth at the extrapolated position \(\tilde{\mathbf{x}}_t = \mathbf{x}_t + \Delta s\, \mathbf{u}_t\) is
\begin{equation}
    \tilde{z}_t = z_t + \Delta s\, \frac{z_{t+1} - z_t}{\|\mathbf{x}_{t+1} - \mathbf{x}_t\|}.
\end{equation}

Given the relatively smooth and predictable characteristics of many lake environments, we can assume that this gradient-based estimation provides a good approximation. We therefore introduce a pseudo-observation $(\tilde{\mathbf{x}}_t, \tilde{z}_t)$, derived from this estimate, to update our GP model (Equation \ref{eq:GP}) and the Proxy with a confidence proxy $C(\tilde{\mathbf{x}}_t) = \alpha$, where \(\alpha < 1\) is a hyperparameter that can be adjusted to reflect the underlying confidence in the estimation. The effectiveness of this method is confirmed by our ablation studies.

\subsection{POMDP Formulation}  
A Partially Observable Markov Decision Process (POMDP) is a tuple $(S, A, O, P, Z, R, \gamma)$,
where $S$ is a set of partially observable \emph{states};
$A$ is a set of \emph{actions};
$Z$ is a set of \emph{observations};
$P$:~$S\times A \rightarrow \Pi(S)$ is the \textit{state-transition model}, mapping to a probability distribution $\Pi(\cdot)$ over states;
$O$:~$S\times A \rightarrow \Pi(Z)$ is the \textit{observation model};
$R$ is the \textit{reward function} and $\gamma \in [0,1)$ is a \textit{discount factor}.
The probability distribution over states $\pazocal{B} = \Pi(S)$, called \emph{belief}, is used to model uncertainty about the true state.
The goal of solving a POMDP, e.g., with RL, is to compute a policy \(\pi : \pazocal{B} \rightarrow A\), to maximize the expected return:  
\begin{equation}    
\mathbb{E}_{a_t \sim \pi} \left[ \sum_{t=0}^{\infty} \gamma^t R(s_t,a_t) \right ].
\end{equation}

\begin{figure}
    \centering    \includegraphics[width=0.9\linewidth]{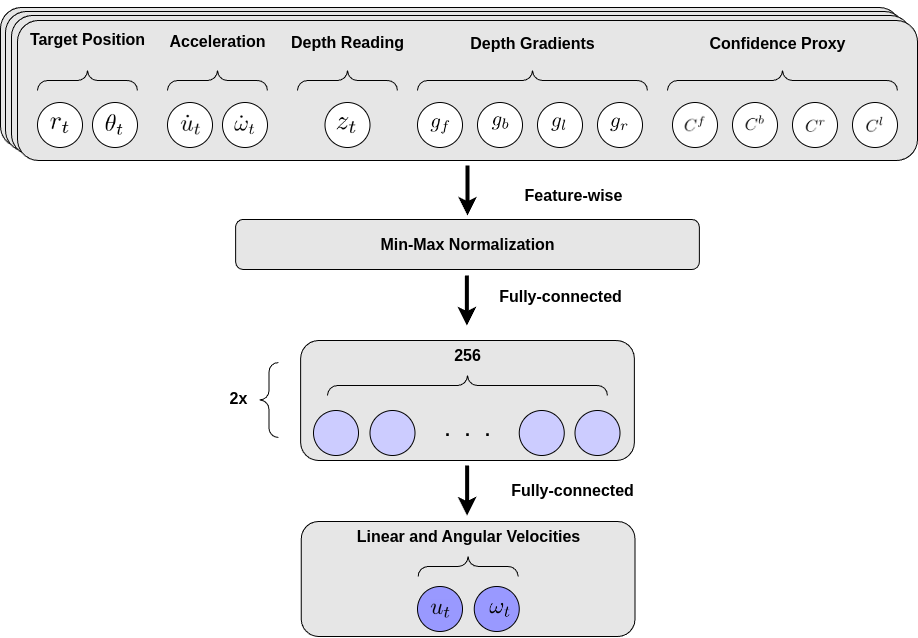}
    \vspace{-5pt}
    \caption{Illustration of the observation vector and network architecture used.}
    \vspace{-10pt}
    \label{fig:final_vs_lstm}
\end{figure}
We now specialize the POMDP formulation to the depth-constrained navigation problem as follows:
\subsubsection{Transition Model}
For practical control applications, a reduced-order 3 DOF model of the ASV dynamics (explained in \ref{method.dynamics}) is typically adopted by constraining pitch (\( q \)), roll (\( p \)), and heave (\( w \)) motions through the assumptions \( p = q = w = 0 \). This yields the simplified horizontal-plane dynamics:
\begin{equation}
\label{eq:dynamic_model}
    m(\dot{u} - rv) = X_{\text{tot}}, \quad 
    m(\dot{v} + ru) = Y_{\text{tot}}, \quad 
    I_z \dot{r} = N_{\text{tot}}
\end{equation}
This simplified formulation retains only surge, sway, and yaw dynamics while eliminating cross-coupling effects from vertical and rotational modes, maintaining sufficient fidelity for surface navigation control objectives \cite{gierusz2016modelling}.

\subsubsection{Action space}
At each time step $t$, the policy outputs a two-dimensional action vector $a_t = [u_t, \omega_t]$ where $u$ and $\omega$ are the linear and angular velocities in the body frame of the ASV, respectively. The action space is bounded within the kinematic limits of the ASV.
% To ensure bounded and physically realizable control actions, we use the \textit{Tanh} activation function at the last layer of the policy network to constrain the action space to $A=[-1,1]^2$, which are then mapped to the actual linear and angular velocities.

\subsubsection{Observation space} 
At each time step \( t \), the agent receives a temporal sequence of observations $\{o_i\}_{t-T}^t$, including key environmental and agent-related variables. Specifically, each observation $o_i$ includes the relative target's position in polar coordinates as \( (r_i, \theta_i) \), where \( r_i \) denotes the radial distance and \( \theta_i \) denotes the angular distance in boat frame; the linear acceleration \( \dot{u}_i \) and angular acceleration \( \dot{\omega}_i \); the previous action \( a_{i-1} \); the current depth reading \( z_i \); the estimated depth gradients and the confidence proxy in four primary directions: forward, backward, right and left. Let $\mathbf{x}^m_i$ be the adjacent position in the main directions to the $\mathbf{x}_i$, where $m \in \{f,b,l,r \}$. For each direction, the gradient is computed as:
    \begin{equation}    
        g^m_i = \frac{\mu_{t+1}(\mathbf{x}^m_i) - z_i}{\mathbf{x}^m_i - \mathbf{x}_i}
    \end{equation}
where \( z_i \) is the current depth reading and \( \mu_{t+1}(\mathbf{x}^m_i) \) is the predictive mean function evaluated at the corresponding position. Additionally, the confidence proxy values $C(\mathbf{x}^m_i)$ where $m \in \{f,b,l,r \}$ represent the uncertainty associated with each gradient estimate. This structured representation enables the policy to interpret depth variations and environmental changes based on partial observations, aiding decision-making.

\begin{figure}
    \centering\includegraphics[width=\linewidth]{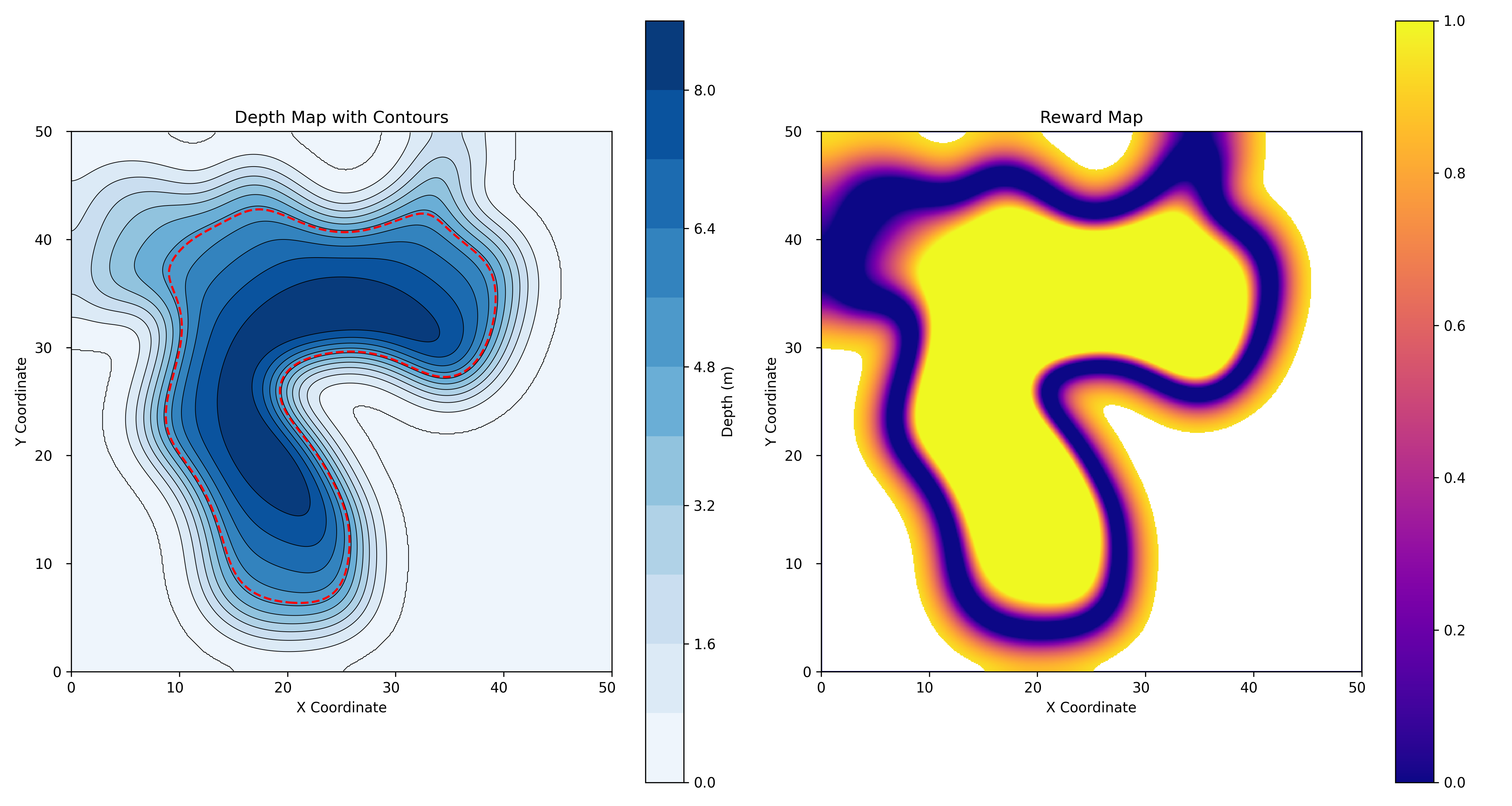}
    \vspace{-20pt}
    \caption{Illustration of $r_{\rm depth}$.}
    \vspace{-10pt}
    \label{fig:r_depth}
\end{figure}
\subsubsection{Reward function}
The Bathymetry-Aware ASV navigation task aims to minimize the distance to the goal while ensuring that the ASV remains clear of unsafe areas. To achieve this objective, we define the following reward function:\begin{equation}\label{eq:reward}
        R(s_t, a_t, s_{t+1}) = 
        \begin{cases} 
        r_{suc} & \text{if }success; \\
        r_{fail} & \text{if }fail; \\
        -\alpha_1 \Delta_{\text{d}} - \alpha_2 r_{\text{back}} - \alpha_3 r_{\text{depth}} & \text{otherwise},
        \end{cases}
\end{equation}
where $\Delta_{d}$ denotes the change in the geodesic distance to the goal from the previous state, $r_{back} = |\min(u_t, 0)|$ quantifies the extent of backward movement and $r_{depth}$ at time is defined by 
\begin{equation}
\label{eq:r_depth}
        r_{\text{depth}} =
        \begin{cases}
        0 & \text{if } \frac{1}{3} L_{\text{d}} \leq z_t < \frac{2}{3} L_{\text{d}} \\
        \frac{3}{L_d} \left( z_t - \frac{2}{3} L_{\text{d}} \right) & \text{if } \frac{2}{3} L_{\text{d}} \leq z_t < L_{\text{d}} \\
        \frac{3}{L_d} \left( z_t - \frac{1}{3} L_{\text{d}} \right) & \text{if } 0 \leq z_t < \frac{1}{3} L_{\text{d}} \\
        \end{cases}
    \end{equation}
where \( z_t \) is the depth measurement at time \( t \) and \( L_{d} \) is the depth limit that defines the unsafe regions. In this formulation, no penalty is applied when the depth lies within the safe range, \([\frac{1}{3} L_{d}, \frac{2}{3} L_{d})\). However, as the depth measurement approaches either the lower or upper bounds (i.e., deviates from the safe interval), the penalty increases linearly. An illustration of this reward function is presented at Figure \ref{fig:r_depth}. This design effectively encourages the agent to maintain a safe operating depth, thereby avoiding proximity to both the shore and excessively deep regions.

In addition to the continuous reward component, discrete rewards are provided for specific events. A large negative
reward $r_{fail} = -100$ is given upon entering to unsafe areas, or timeout event, terminating the episode.
A large positive reward $r_{suc}=200$ is given when the target is reached.

%% file: sections/4.simulation_experiments.tex
\section{Experiments}
% For successful ASV navigation, a single sonar reading is insufficient for our feature extractor to effectively guide the action network to safely navigate the ASV. To provide richer information, we employ GP to estimate the depth gradient in the local area of the boat.

We design our experiments to answer the following research questions: 
\begin{enumerate}
    \item[RQ1] How does the success rate of our method compare to that of a privileged model with full access to the lake's depth map?
    \item[RQ2] What is the impact of GP regression with respect to the single SBES information from the depth sensor?
    \item[RQ3] How does our model compare to a safe RL approach \cite{achiam2017constrained}, where the goal is to find an optimal policy under the constraint to keep $r_{\rm depth} = 0$ (i.e., the ASV shall remain in the safe zone)?
    \item[RQ4] How does our online GP-based estimation of the local depth map compare to a pre-trained LSTM network?
    \item[RQ5] How well does our methodology generalize out of the training dataset, both in simulation and in a real depth-constrained navigation task?
\end{enumerate}
% $(i)$  $(ii)$ Given that our model lacks critical information, can it learn to operate as efficiently as the privileged model? $(iii)$ In our safety-critical task, does Safe RL outperform standard RL? $(iv)$ What are the effects of using a learning-based depth/depth gradient estimation, and how well does it generalize to the environment?
We begin by describing the implementation details and the training setup. Next, we assess the performance of our approach in simulation, followed by an evaluation of the learned policy in real-world experiments.

\subsection{Implementation and training details}
We generate 120 depth maps in simulation for training (Figure \ref{fig:depth_maps}), randomizing the shoreline and depth values with radial gradients, noise and filtering techniques. The maps are sized $50 \times 50$ \si{m} with a depth resolution of \SI{10}{cm}$^2$. We used the dynamic model of the ASV and implement a vectorized OpenAI gym environment, which allows for simulating multiple ASV's in parallel, achieving approximately 2000 FPS on a single GeForce 4070 GPU. The RL environment is implemented using the Stable-Baselines3 library \cite{stable-baselines3}, and the safe environment is constructed with the OmniSafe library \cite{ji2024omnisafe}. For our proposed method, we employ a two-layer multilayer perceptron (MLP) with 256 units per layer. We train our agent using Soft Actor-Critic (SAC) \cite{haarnoja2018soft}, which is a popular off-policy reinforcement learning algorithm that maximizes reward while optimizing for entropy, encouraging exploration and robust policies. For safe-RL, we used the Lagrangian version of SAC (SACLag) \cite{ray2019benchmarking}. For the kernel function, we used the well-known Radial Basis Function (RBF) kernel \cite{williams2006gaussian} with a length-scale set to 3.0 empirically. We trained all the models for 5 million steps and for 5 random seeds. We evaluated approaches using five metrics: Success Rate (SR) measuring task completion percentage, Efficiency Score (ES) calculated as success rate divided by episodic length, Depth Break (DB) indicating episodes terminated by depth constraint violations, Velocity Smoothness (VS) quantifying velocity change smoothness, and Heading Smoothness (HS) measuring heading direction change rates~\cite{karwowski2023quantitative}.

\subsection{Simulation Results}
We first evaluate our proposed approach in simulation, against several baselines to comprehensively assess the performance:

\begin{enumerate}
    \item \textbf{Privileged model:} This model has privileged access to true depth gradients, bypassing estimation and serving as a performance upper bound.
    \item \textbf{Model w/o GP:} In this variant, the GP information is omitted from the state space.
    \item \textbf{Safe RL:} In this model, the cost is computed from \( r_{\text{depth}} \) in Equation \eqref{eq:r_depth}, omitting it from the reward function \eqref{eq:reward}.
\end{enumerate}

\begin{figure}
    \centering
    \includegraphics[width=\linewidth]{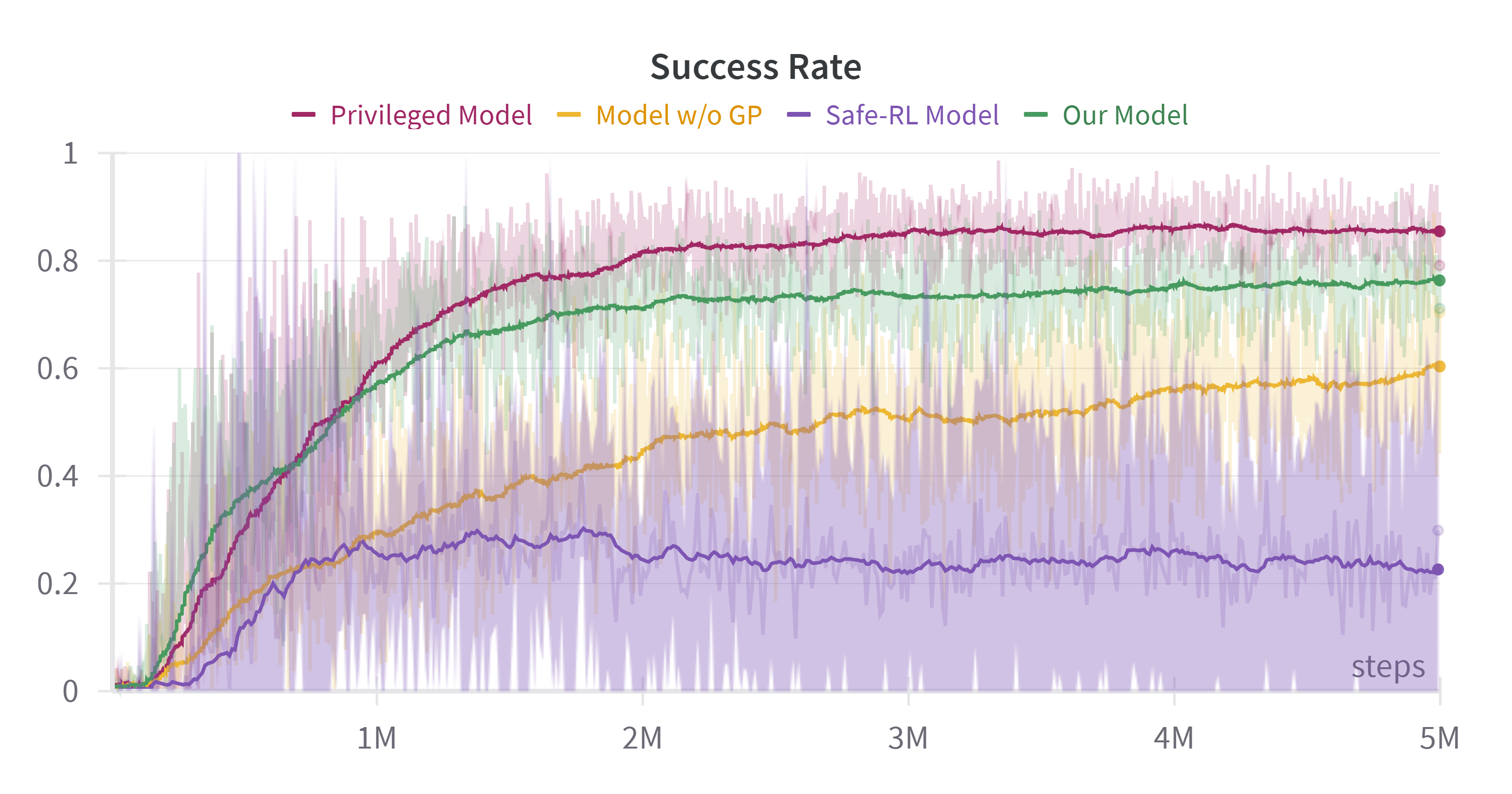}
    \vspace{-20pt}
    \caption{Mean success rate over 5 million training steps, averaged across 5 random seeds for each approach.}
    \vspace{-10pt}
    \label{fig:all_models}
\end{figure}
\begin{figure*}
    \centering
    \includegraphics[width=0.7\linewidth]{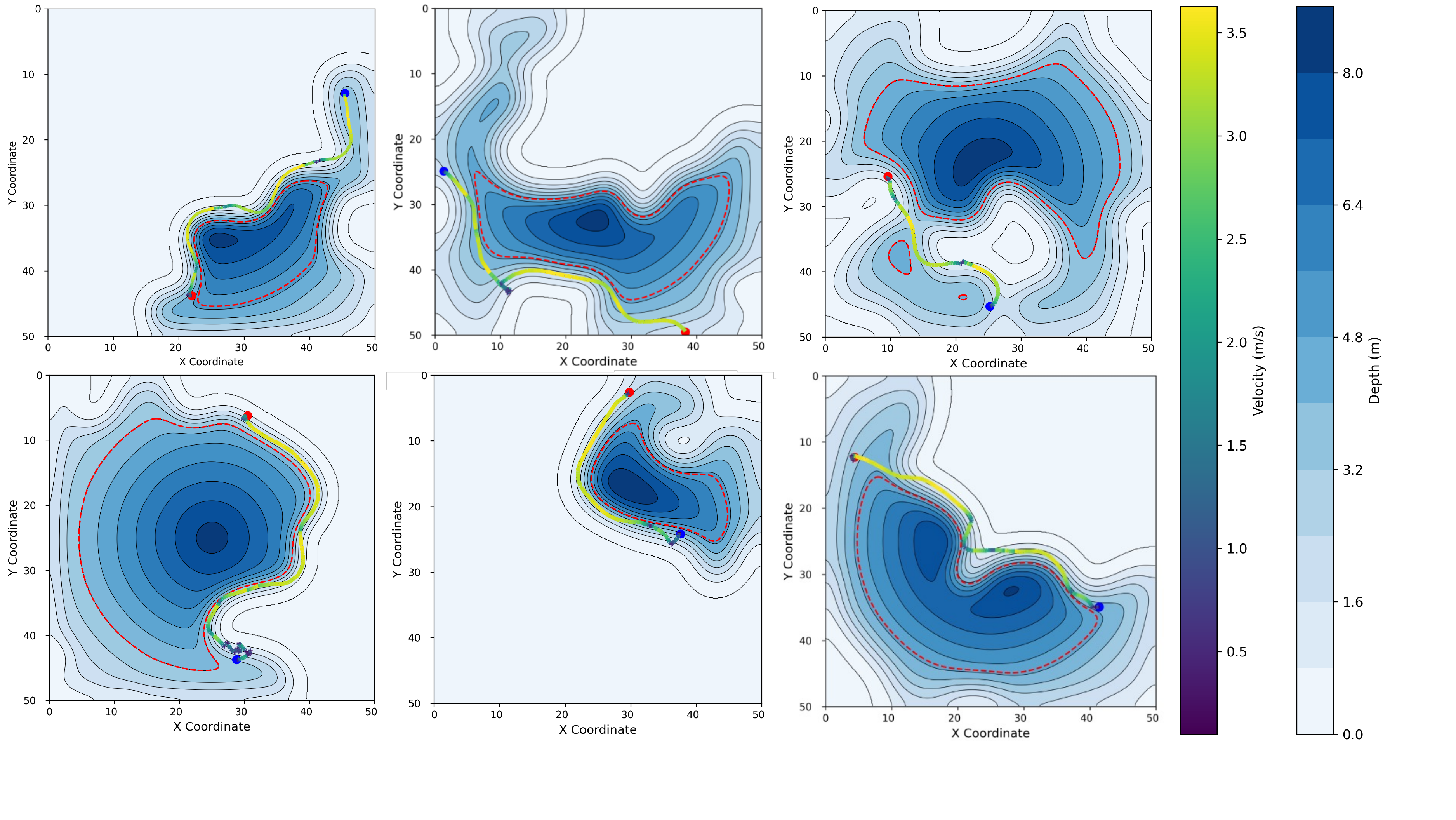}
    \vspace{-25pt}
    \caption{ASV navigation trajectories showing multiple path strategies from starting points (blue) to targets (red), with color gradients representing velocity variations across different scenarios.}
    \vspace{-10pt}
    \label{fig:trajectories}
\end{figure*}
\begin{figure}
    \centering
    \includegraphics[width=\linewidth]{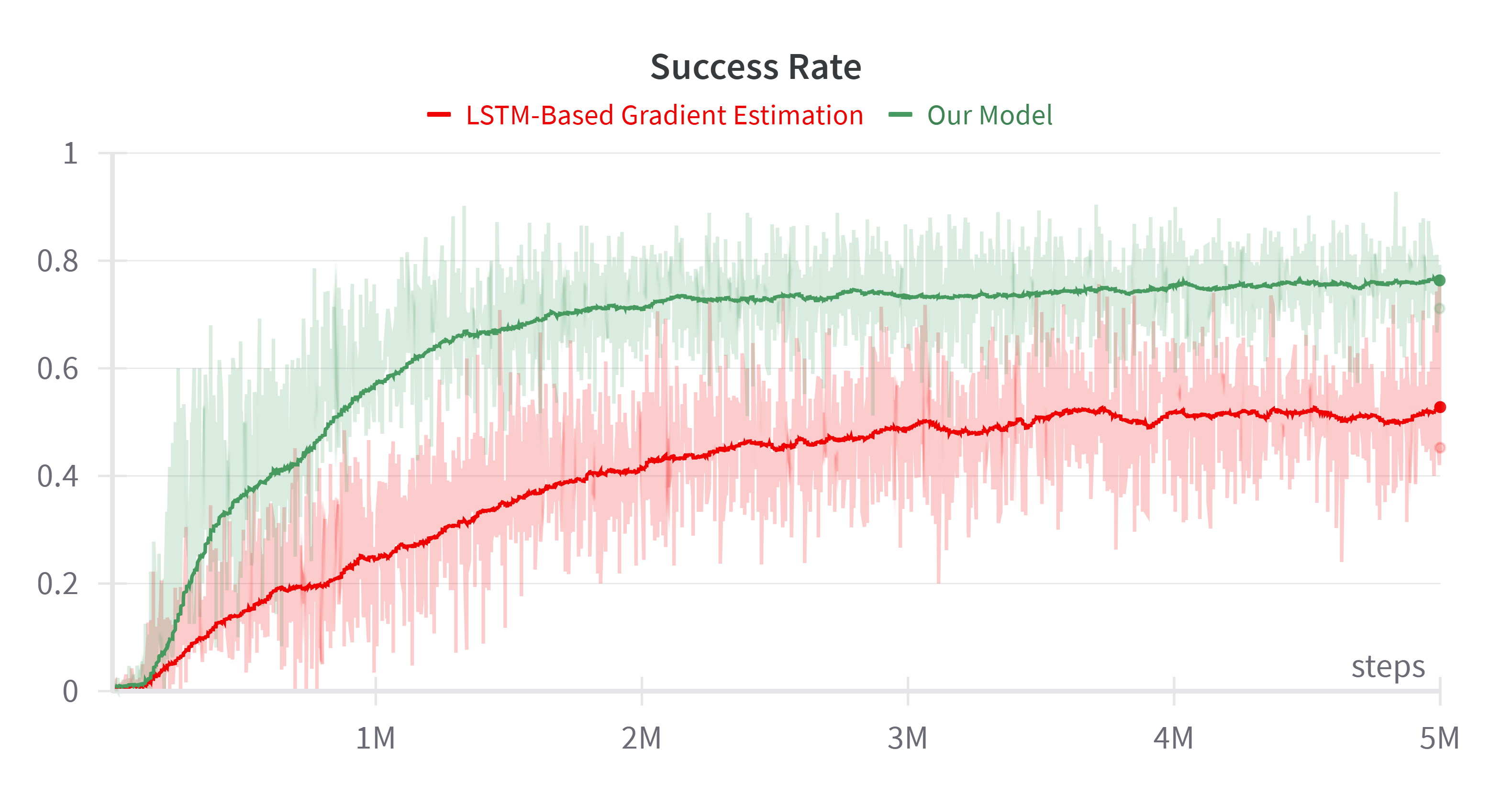}
    \vspace{-20pt}
    \caption{Mean success rate over 5 million training steps, averaged across 5 random seeds for each approach.}
    \vspace{-10pt}
    \label{fig:final_vs_lstm}
\end{figure}
Figure \ref{fig:all_models} shows the success rate (i.e., percentage of episodes where the policy reaches the target while keeping the depth constraint) along the learning curve of the different algorithms on the training maps. Our model (green) achieves 76.33\% on average, with limited variance (hence being more robust). 
This is slightly worse than the performance achieved by the privileged model (red), which attains 85.45\% success rate on average, showing that our GP-enhanced RL architecture well approximates full environmental awareness, requiring less sensory information (RQ1).
In contrast, the model without GP information in the state space (yellow) achieves 60.34\% success rate on average with larger variance, thus highlighting the key contribution of GP regression (RQ2). Similarly, the safe RL approach (purple) gets the worst performance (22.6\% on average), possibly because the constraint imposed by the cost function in SACLag restricts exploration and learning efficiency, leading to suboptimal policy updates and reduced overall performance. Additionally, it exhibits high variance, which suggests instability in learning (RQ3).

To address RQ4, We also compared the success rate of our approach against an LSTM-based gradient estimation method. This method employs a Long Short-Term Memory (LSTM) network trained on 65k data from pre-trained models to directly estimate depth gradients. During RL training, the LSTM parameters remain frozen, and its performance is evaluated against our proposed model, which integrates state estimation. As shown in Fig. \ref{fig:final_vs_lstm}, learning-based gradient estimation struggles due to distribution shifts in the LSTM model, leading to a 23.55\% performance drop, from 76.33\% (our model) to 52.78\%.

% For the final evaluation, we tested all models on the depth maps encountered during training over 1,000 episodes, with the results summarized in Table \ref{tab:all_results}. The Efficiency Score is calculated as the Success Rate divided by the Mean Episode Length, providing a more reliable measure of each model's efficiency.

\begin{table}
\caption{Performance comparison of different models on the navigation task on 100 unseen depth maps for 1000 episodes.}
\vspace{-5pt}
\label{tab:all_results_unseen}
\centering
\resizebox{0.5\textwidth}{!}{
\begin{tabular}{c|ccccc}
\hline
Method           & SR (\%)↑         & ES ↑ & MDB (\%)↓ & VS ↓ & HS ↓\\ \hline
Privileged & 85.78 {\scriptsize(±1.54)} & 0.781 {\scriptsize(±0.15)}         & 5.66 {\scriptsize(±1.16)}     & 69.71 {\scriptsize(±6.78)} & 24.70 {\scriptsize(±15.06)} \\ \hline
 w/o GP     & 54.52 {\scriptsize(±2.42)} & 0.222 {\scriptsize(±0.14)}         & \textbf{10.98} {\scriptsize(±2.12)}   & \textbf{36.58} {\scriptsize(±7.20)} & \textbf{8.73} {\scriptsize(±4.72)} \\
Safe-RL     & 23.98 {\scriptsize(±20.73)} & \underline{0.325} {\scriptsize(±1.15)}         & 58 {\scriptsize(±16.83)} & 88.96 {\scriptsize(±8.95)} & 68.98 {\scriptsize(±12.50)} \\
LSTM-based       & \underline{56.08} {\scriptsize(±2.21)} & 0.220 {\scriptsize(±0.04)}         &  16.62 {\scriptsize(±4.25)} & \underline{44.68} {\scriptsize(±4.12)}  & 25.44 {\scriptsize(±10.19)} \\
Our Model        & \textbf{76.26} {\scriptsize(±2.44)} & \textbf{0.745} {\scriptsize(±0.28)}          & \underline{12.16} {\scriptsize(±1.64)}   & 56.64 {\scriptsize(±11.02)} & \underline{22.21} {\scriptsize(±11.52)}\\ \hline
\end{tabular}
}
\vspace{-10pt}
\end{table}

To evaluate generalization (RQ5), we tested the best policies on 100 unseen depth maps, running 1{,}000 episodes per model (see Table~\ref{tab:all_results_unseen}). Results align with findings from RQ1--4. Our model performs closest to the privileged one (76.26\% vs. 86.78\% success), significantly outperforming the w/o GP baseline (54.52\%) and LSTM (56.08\%), highlighting the importance of online GP regression over static neural models.
We also assessed the \emph{efficiency score}, where our method again ranks highest (0.745), close to the privileged upper bound (0.781). For \emph{mean depth break} (percentage of episodes violating depth constraints), our approach ranks second (12.16\%), slightly behind the no-GP model (10.98\%) but with significantly higher success and efficiency.

% Overall, our approach successfully recovered 68.1\% of the performance gap on seen maps and 69.5\% on unseen maps during evaluation. Additionally, our model achieved the second-best performance in terms of Mean Depth Break. The Model w/o GP outperformed it in this metric because it struggled to learn a successful navigation strategy. Instead of prioritizing goal-reaching, it optimized \( r_{depth} \) to avoid unsafe areas and minimize large negative rewards.

\begin{table*}
\centering
\caption{Ablation study results showing the impact of different components. The last line represents our full model. Row 2 uses techniques proposed by~\cite{nguyen2008local}, while row 4 combines techniques from both~\cite{nguyen2008local} and~\cite{snelson2005sparse}.}
\vspace{-5pt}
\label{tab:ablation_results}
\begin{tabular}{c|cccc|ccccc}
&GP&Gradient-Based Extrapolation& Variance & Proxy & SR (\%)↑& ES ↑& MDB (\%)↓ & VS ↓ & HS ↓  \\ \hline
1&\ding{55}&\ding{55}&\ding{55}&\ding{55}&54.52 {\scriptsize(±2.42)} & 0.222 {\scriptsize(±0.14)}         & \textbf{10.98} {\scriptsize(±2.12)}   & \textbf{36.58} {\scriptsize(±7.20)} & \textbf{8.73} {\scriptsize(±4.72)}\\ 
2&\ding{51}&\ding{55}&\ding{55}&\ding{55}&62.4 {\scriptsize(±1.97)} & 0.289 {\scriptsize(±0.03)}&24.07 {\scriptsize(±6.18)} & \underline{51.66} {\scriptsize(±12.99)} & 28.10 {\scriptsize(±12.70)}\\
%3&\ding{51}&\ding{55}&\ding{51}&\ding{55}& 42.22 {\scriptsize(±3.55)}& 0.107 {\scriptsize(±0.05)} & 8.80 {\scriptsize(±2.09)} & 46.64 {\scriptsize(±3.40)} & 23.03 {\scriptsize(±2.98)}      \\ 
3&\ding{51}&\ding{51}&\ding{55}&\ding{55}& \underline{72.76} {\scriptsize(±4.63)}& 0.488 {\scriptsize(±0.10)} & 18.78 {\scriptsize(±4.14)} & 51.78 {\scriptsize(±15.41)} & \underline{21.51} {\scriptsize(±6.55)}      \\ 
4&\ding{51}&\ding{51}&\ding{51}&\ding{55}& 69.9 {\scriptsize(±2.52)} & \underline{0.693} {\scriptsize(±0.30)}&16.16 {\scriptsize(±1.33)} & 60.99 {\scriptsize(±7.41)} & 23.14 {\scriptsize(±22.49)}\\ 
5&\ding{51}&\ding{51}&\ding{55}&\ding{51}& \textbf{76.26} {\scriptsize(±2.44)} & \textbf{0.745} {\scriptsize(±0.28)}          & \underline{12.16} {\scriptsize(±1.64)}   & 56.64 {\scriptsize(±11.02)} & 22.21 {\scriptsize(±11.52)}       
\end{tabular}
\vspace{-10pt}
\end{table*}

In our analysis, an interesting pattern emerged when comparing model performance. Despite our model achieving a success rate approximately 10\% lower than the privileged model, the efficiency scores between the two were similar. This efficiency score can be attributed to our model's shorter average episode length. The privileged model exhibited a distinctive movement pattern characterized by rapid acceleration followed by abrupt deceleration when detecting depth gradients indicative of unsafe areas. This resulted in a sinusoidal trajectory along safe regions. Conversely, our model operated with inherent uncertainty regarding depth readings and estimations, leading to more conservative movement strategies. It's important to note that this smoother trajectory is not an inherent advantage of our approach, but rather a side effect of the uncertainty in our model's perception system. With appropriate reward functions designed to optimize trajectory smoothness, the privileged model could potentially achieve both higher success rates and smoother navigation. However, since neither model was explicitly optimized for smoothness in our experimental setup, our model's cautious approach manifested as slower but more consistent progression with fewer directional fluctuations compared to the privileged model. This observation is further validated by our study over VS and HS in Table~\ref{tab:all_results_unseen}, where the privileged model used more linear and angular velocity changes.
We finally validate our methodology on the real ASV shown in Figure \ref{fig:real_asv}, to assess the capability of our RL methodology to overcome the notably challenging sim-to-real gap. The attached video to this submission shows the attained performance.

%\begin{table}[]
%\caption{AJAB}
%\label{tab:my-table}
%\begin{tabular}{c|ccc}
% Method           & Success Rate  & Mean Episode Length & Mean Depth Break \\ \hline
% Privileged Model & 87.33 +- 3.43 & 125.88+- 11.87      & 8.4+- 2.97       \\
% Model w/o GP     & 58.66+- 8.25  & 326.63+- 118.89     & 11.32+- 6.18     \\
% Safe-RL Model    & 14.86 +- 9.59 & 83.26 +- 11.76      & 62.66 +- 11.64   \\
% Our Model        & 83.33 +- 5.16 & 102.73+- 17.68      & 11.48+- 2.90    
% \end{tabular}
% \end{table}

% \begin{table}
% \caption{Performance comparison of different models on the navigation task on seen depth maps.}
% \label{tab:all_results}
% \centering
% \begin{tabular}{c|ccc}
% \hline
% Method           & Success Rate (\%)          & Efficiency Score & Mean Depth Break \\ \hline
% Privileged Model & 84.32 {\scriptsize(±1.93)} & 0.728 {\scriptsize(±0.19)}         & 5.94 {\scriptsize(±1.63)}      \\ \hline
% Model w/o GP     & 52.4 {\scriptsize(±2.86)} & 0.20 {\scriptsize(±0.12)}         & \textbf{11.58} {\scriptsize(±2.64)}    \\
% Safe-RL Model    & 22.26 {\scriptsize(±18.59)} & \underline{0.297} {\scriptsize(±1.34)}         & 58.82 {\scriptsize(±14.47)}   \\
% LSTM-based       & \underline{54.04} {\scriptsize(±2.72)} & 0.204 {\scriptsize(±0.04)}         &  17.28 {\scriptsize(±5.64)} \\
% Our Model        & \textbf{74.14} {\scriptsize(±3.94)} & \textbf{0.698} {\scriptsize(±0.43)}          & \underline{13.84} {\scriptsize(±2.33)}    \\ \hline
% \end{tabular}
% \end{table}

\subsection{Ablation Study}

%To assess the contribution of each component in our model, we conducted an ablation study by systematically removing key elements and evaluating their impact on performance. We tested variations of our model over 1,000 episodes. The results, summarized in Table \ref{tab:ablation_results}, reveal that removing the Gaussian Process module significantly reduces performance, confirming its role in improving depth-aware navigation. Furthermore, Future Update component significantly enhance model performance. Enabling GP alone (row 2) improves the baseline, and its combination with Future Update (row 3) further boosts the Success Rate and Efficiency Score. As mentioned earlier, using Variance may result in overconfidence estimations. This was confirmed by adding the Variance components (row 4), which make the success rate degrade by 2.52\% and the model violated depth constraint 1.8\% more. Notably, replacing Variance with the Proxy module (row 5) results in the highest Success Rate (74.14\%) and Efficiency Score (0.698), while maintaining a moderate Mean Depth Break. Results of row 4 and 5 confirms that having information about uncertainty helps the model to operate reach the target in shorter episode lengths, highlighting the role of knowing uncertainty. These results affirm that GP, Future Update, and Proxy are critical for optimal depth-aware navigation.

We conducted an ablation study over 1,000 episodes in Table \ref{tab:ablation_results}, to highlight the independent contribution of 3 key items in our methodology: GP estimation, Gradient-Based Extrapolation (explained in \ref{future_update}) and proxy (Equation \ref{eq:proxy}). Removing the GP module significantly reduced performance, confirming its role in depth-aware navigation. Enabling GP (row 2) \cite{nguyen2008local} improved the baseline, while combining it with Gradient-Based Extrapolation \cite{snelson2005sparse} (row 3) further boosted the Success Rate and Efficiency Score. Adding the Variance component (row 4) led to a 2.52\% drop in Success Rate and a 1.8\% increase in depth violations, likely due to overconfident estimations. Replacing Variance with the Proxy module (row 5) achieved the highest Success Rate (74.14\%) and Efficiency Score (0.698) with moderate depth constraints. These results affirm that GP, Gradient-Based Extrapolation, and Proxy are critical for optimal performance.

\subsection{Real-World ASV Navigation}

To evaluate the transferability of our policy from simulation to the real world, we deployed it on a physical ASV built using off-the-shelf components, including a 3D printed frame, two underwater thrusters, one SBES sensor, and a Raspberry Pi 3. The ASV was tasked with reaching randomly generated target points in a real aquatic environment.
We conducted five experiments, with the ASV successfully reaching the target in four out of five runs. The only failure was due to a depth constraint violation, caused by fluctuations in the depth sensor's sampling frequency. The efficiency score (ES) was 0.530; vertical and heading stability metrics (VS and HS) were not computed due to the lack of an onboard IMU.
While the policy showed generally reliable performance, we observed fluctuations in control frequency stemming from the Raspberry Pi’s limited processing power, occasionally leading to degraded control. Communication delays, which were not modeled during training, also contributed to the sim-to-real performance gap.

%% file: sections/5.Conclusion.tex
\section{CONCLUSION}

In this work, we presented a reinforcement learning-based approach for ASV navigation under depth constraints, enabling safe and efficient operation in shallow-water environments with minimal sensor information. Our method integrates Gaussian Process regression to estimate a bathymetric depth map from sparse sonar readings, improving environmental awareness and decision-making.

We validated our approach through extensive simulation experiments and demonstrated its transferability to real-world aquatic scenarios. The policy achieved reliable real-world performance despite limited onboard computation and unmodeled environmental factors, highlighting its robustness and practical potential.

A key limitation lies in the reliance on smooth depth estimations from GP regression, which may struggle with abrupt terrain changes or unmodeled obstacles such as submerged rocks. To address this, future work will also explore enhanced sensing strategies (e.g., additional sonar units) and investigate mapless navigation approaches, where the ASV operates without an explicit depth map.
To improve real-world performance, especially under limited hardware and communication constraints, future work will explore domain randomization and adaptation techniques to enhance robustness and reduce the sim-to-real gap. 